%% file: paper.tex
\documentclass[conference]{IEEEtran}
\IEEEoverridecommandlockouts

\usepackage{cite}
\usepackage{amsmath} 
\usepackage{amssymb}  

\usepackage{graphicx}
\usepackage{capt-of} 
\usepackage{overpic} 
\usepackage[export]{adjustbox} 
\usepackage{textcomp} 
\usepackage{tabularx}
\usepackage{multirow}
\usepackage{booktabs}
\usepackage{arydshln}
\usepackage{adjustbox} 
\usepackage{pgfplots} 
\usepackage{siunitx}
\usepackage[hyperfootnotes=false]{hyperref}
\usepackage{balance}

\usepackage{algorithm}
\usepackage{algpseudocode}
\algblock{Input}{EndInput}
\algnotext{EndInput}
\algblock{Output}{EndOutput}
\algnotext{EndOutput}
\newcommand{\Desc}[2]{\State \makebox[5em][l]{#1}#2}

\makeatletter
\let\OldStatex\Statex
\renewcommand{\Statex}[1][3]{%
	\setlength\@tempdima{\algorithmicindent}%
	\OldStatex\hskip\dimexpr#1\@tempdima\relax}
\makeatother

\usepackage[caption=false, font=scriptsize]{subfig} 

\newcommand\copyrighttext{\footnotesize \textcopyright~2023 IEEE. Personal use of this material is permitted. Permission from IEEE must be obtained for all other uses, in any current or future media, including reprinting/republishing this material for advertising or promotional purposes, creating new collective works, for resale or redistribution to servers or lists, or reuse of any copyrighted component of this work in other works.
}%

\newcommand\copyrightnotice{%
	\begin{tikzpicture}[remember picture,overlay]
	\node[anchor=south,xshift=0pt,yshift=14pt] at (current page.south) {\fbox{\parbox{\dimexpr\textwidth-\fboxsep-\fboxrule\relax}{\copyrighttext}}};
	\end{tikzpicture}%
}

\begin{document}
\title{LMR: Lane Distance-Based Metric for Trajectory Prediction
	\thanks{The research leading to these results is funded by the German Federal Ministry for Economic Affairs and Climate Action within the project "KI Delta Learning" (F\"orderkennzeichen 19A19013A). The authors would like to thank the consortium for the successful cooperation.}
	\thanks{$^{1}$Source code: \url{https://github.com/schmidt-ju/lane-miss-rate}}%
}

\author{\IEEEauthorblockN{Julian Schmidt\IEEEauthorrefmark{1}\IEEEauthorrefmark{2}, Thomas Monninger\IEEEauthorrefmark{3}, Julian Jordan\IEEEauthorrefmark{1} and Klaus Dietmayer\IEEEauthorrefmark{2}}
	\IEEEauthorblockA{\IEEEauthorrefmark{1}Mercedes-Benz AG, Research \& Development, Stuttgart, Germany\\
		Email: julian.sj.schmidt@mercedes-benz.com}
	\IEEEauthorblockA{\IEEEauthorrefmark{2}Ulm University, Institute of Measurement, Control and Microtechnology, Ulm, Germany}
	\IEEEauthorblockA{\IEEEauthorrefmark{3}Mercedes-Benz Research \& Development North America, Sunnyvale, CA, USA}}

\maketitle

\begin{abstract}
The development of approaches for trajectory prediction requires metrics to validate and compare their performance.
Currently established metrics are based on Euclidean distance, which means that errors are weighted equally in all directions.
Euclidean metrics are insufficient for structured environments like roads, since they do not properly capture the agent's intent relative to the underlying lane.
In order to provide a reasonable assessment of trajectory prediction approaches with regard to the downstream planning task, we propose a new metric that is lane distance-based: Lane Miss Rate (LMR).
For the calculation of LMR, the ground-truth and predicted endpoints are assigned to lane segments, more precisely their centerlines.
Measured by the distance along the lane segments, predictions that are within a certain threshold distance to the ground-truth count as hits, otherwise they count as misses.
LMR is then defined as the ratio of sequences that yield a miss.
Our results on three state-of-the-art trajectory prediction models show that LMR preserves the order of Euclidean distance-based metrics.
In contrast to the Euclidean Miss Rate, qualitative results show that LMR yields misses for sequences where predictions are located on wrong lanes.
Hits on the other hand result for sequences where predictions are located on the correct lane.
This means that LMR implicitly weights Euclidean error relative to the lane and goes into the direction of capturing intents of traffic agents.
The source code of LMR for Argoverse~$2$ is publicly available$^1$.
\end{abstract}


\section{Introduction}
\copyrightnotice 
In order to plan their future motion, autonomous vehicles are required to predict the motion of surrounding agents.
Recently, there has been a large amount of interest in learning-based trajectory prediction, resulting in a myriad of different approaches, e.g., predicting trajectories based on a vectorized~\cite{Gao2020, Liang2020, Schmidt2022} or rasterized ~\cite{PhanMinh2020, Djuric2020} scene representation.

Given this variety of approaches, also methods that allow a quantitative comparison are required.
For this purpose, prediction metrics, such as Average Displacement Error (ADE), Final Displacement Error (FDE) and Miss Rate (MR), exist.

\begin{figure}[!t]
	\centering
	\subfloat[Euclidean distance-based Miss Rate]{%
		\includegraphics[width=1\linewidth, page=2, trim=0cm 0cm 0 0, clip]{figures/motivation.pdf}
		\label{subfig:euclidean_distance-based}}
	\\
	\subfloat[Lane distance-based Miss Rate]{%
		\includegraphics[width=1\linewidth, page=1, trim=0cm 1.45cm 0 0, clip]{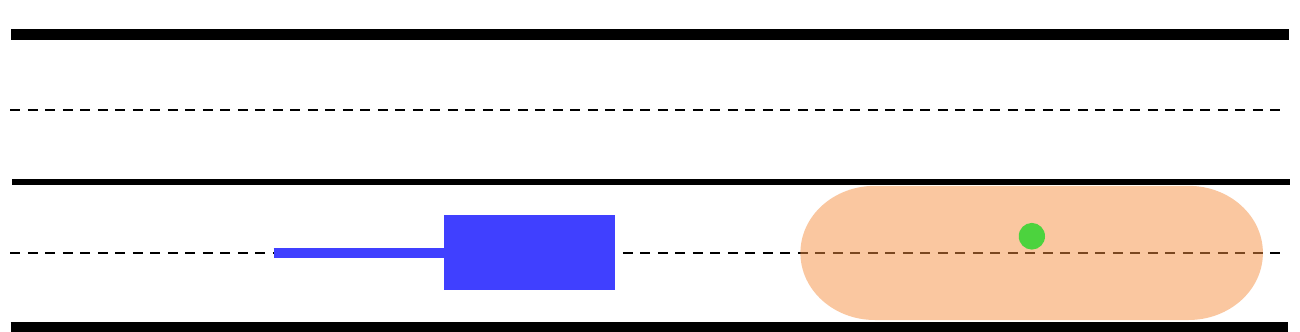}
		\label{subfig:lane_distance-based}}
	\caption{Overview of the established \protect\subref{subfig:euclidean_distance-based} Euclidean distance-based Miss Rate and our \protect\subref{subfig:lane_distance-based} lane distance-based Miss Rate: The green point indicates the ground-truth endpoint. Predictions within the orange area result in a hit and vice versa. Lane centerlines are illustrated with dashed lines.}
	\label{fig:motivation}
\end{figure}

These currently established prediction metrics are based on the Euclidean distance between ground-truth and predictions.
We argue that measuring prediction performance solely by means of spatial distance is insufficient since it does not account for different effects of a displacement at different directions, e.g., due to variations in lane geometry and traffic regulations.
As an example, in some cases, a Euclidean error in longitudinal direction is considerably less severe than the same Euclidean error in lateral distance, which might yield a prediction that is located on the oncoming opposing lane.
Under consideration of the downstream planning task, this means that the same Euclidean error can yield to significantly different results.
Hence, Euclidean error has an asymmetrical effect on the downstream planning task.

In this paper, we propose a lane distance-based metric for the evaluation of trajectory prediction models.
For the calculation of this metric, ground-truth and predicted endpoints are assigned to lane segments.
Measured by the distance along the lane segments, predictions that are within a certain threshold distance to the ground-truth count as hits and predictions that are outside that threshold distance count as misses.

Fig.~\ref{fig:motivation} illustrates how this compares to the already existing Euclidean distance-based MR.
The \subref{subfig:euclidean_distance-based} Euclidean distance-based MR yields hits even for predictions on opposing lanes and outside of the drivable space.
Our \subref{subfig:lane_distance-based} lane distance-based MR, however, aims to consider asymmetrical effects by only yielding hits for predictions that are assigned to the same lane segment as the ground-truth.
This way, our metric goes into the direction of capturing intents of traffic agents.

In summary, our main contributions are:
\begin{itemize}
	\item We propose Lane Miss Rate (LMR), a novel lane distance-based metric for the evaluation of trajectory prediction models.
	\item We extensively compare different state-of-the-art trajectory prediction models on already established metrics and on our newly introduced LMR.
	\item We provide publicly available source code of LMR for Argoverse~$2$ and therefore contribute to the field of trajectory prediction evaluation.
\end{itemize}

\section{Related Work}
This section describes related work regarding the evaluation of trajectory prediction models.
Existing methods are grouped into accuracy-based metrics, quality metrics and planning-based metrics.

\subsection{Accuracy-Based Metrics} \label{subsec:accuracy-based_metrics}
Accuracy-based metrics are mainly Euclidean distance-based and are the most established metrics used for the evaluation of trajectory prediction models~\cite{Chang2019, Ettinger2021, Wilson2021}.
The minimum Average Displacement Error (minADE) and minimum Final Displacement Error (minFDE) are metrics introduced in early work related to learning-based trajectory prediction~\cite{Lee2017, Gupta2018, Rhinehart2018, Rhinehart2019, Chang2019}.
minFDE is the Euclidean distance between the ground-truth endpoint and the predicted endpoint.
For multi-modal predictions, only the prediction with the shortest distance is considered.
minADE is the average Euclidean distance between the ground-truth and the predicted trajectory, with multi-modal predictions being evaluated with the same trajectory used for the calculation of the minFDE.
For multi-modal predictions with probability estimates per mode, brier-minADE and brier-minFDE are applicable~\cite{Brier1950}.
They correspond to the term $(1-p)^2$ being added to the minADE and minFDE value.
The term considers the probability $p$ of the mode selected for the minADE and minFDE calculation.

Miss Rate (MR) is the ratio of sequences, where the Euclidean prediction error is larger than a defined threshold~\cite{Lee2017, Yeh2019}.
Most commonly, the prediction error refers to the Euclidean endpoint error and the threshold is a predefined distance, e.g., \SI{2}{m}~\cite{Chang2019, Wilson2021}.
For multi-modal predictions, all modes must be above the threshold, in order to yield a miss.
Using separate Euclidean thresholds for longitudinal and lateral directions that are additionally dependent on time and velocity is also possible~\cite{Ettinger2021}.

Other noteworthy Euclidean distance-based metrics are mean Average Precision (mAP)~\cite{Ettinger2021}, which is based on MR, and Negative Log Likelihood (NLL)~\cite{Ivanovic2019}, which additionally considers positional uncertainty in predictions.

Our proposed metric is also accuracy-based.
However, in contrast to already exiting accuracy-based metrics, it is dependent on the lane distance instead of pure Euclidean distance.

\subsection{Quality Metrics}
Alternatively, it is possible to measure the quality of a predicted trajectory without considering the ground-truth.
Turn Rate Infeasibility (TRI) is used to measure the ratio of predictions that are kinematically infeasible in terms of the predicted turn radius~\cite{Cui2020}.
Off-road rate and Drivable Area Compliance (DAC) measure the quality of predictions by determining whether a prediction lies within the drivable space given by a map or not~\cite{Chang2019}.

Quality metrics by no means replace existing distance-based metrics.
They only serve as an additional criterion to check for certain requirements (e.g., kinematically feasible predictions), which is why they are not well established.

\subsection{Planning-Based Metrics}
Instead of directly quantifying prediction performance, it is also possible to measure the performance of a planner that utilizes these predictions.
Planning KL Divergence (PKL) is a metric that allows for such a comparison~\cite{Philion2020}.
PKL is used to compare the output of a planner dependent on different object detectors to the output of the same planner using ground-truth objects.
Instead of comparing the planner performance dependent on different object detectors, this principle would also allow for the comparison of the planner performance dependent on different prediction models.
Downside is that prediction performance is not measured in an isolated way.

An alternative approach is to only work with a planner-agnostic cost function for planning~\cite{Ivanovic2021_ARXIV}.
Relying on continuous inverse optimal control to learn weights of the cost function from a large-scale dataset makes this approach dependent on the distribution of the dataset used for learning.

While these planning-based metrics aim to not suffer from the asymmetries of Euclidean distance-based metrics, we argue that simplicity is still an important criterion for the applicability of metrics.
Using planners or planner cost functions is complex to implement and introduces additional dependencies on the planner or the cost function, which is why these metrics are not well established.

\section{Problem Definition}
We define the ground-truth future trajectory of an agent as $\mathbf{T}_\mathrm{gt} = \{ \boldsymbol{\tau}^t \}^{T_f}_{t=1}$, with $T_f$ being the prediction horizon.
Predictions of a trajectory prediction model are denoted as $\mathbf{T}_{\mathrm{pred}} = \{ \mathbf{T}_{\mathrm{pred}, i} \}^k_{i=1}$, with $\mathbf{T}_{\mathrm{pred}, i} = \{ \boldsymbol{\tilde{\tau}}^t_i \}^{T_f}_{t=1}$.
The index $i \in 1, \dots ,k$ considers the multi-modality of future motion by including $k$ predicted modes.

The evaluation of trajectory prediction models comes down to finding a measure of quality for $\mathbf{T}_{\mathrm{pred}}$, possibly depending on $\mathbf{T}_\mathrm{gt}$.

\section{Method}
This section describes LMR, our lane distance-based trajectory prediction evaluation metric.
\input{algorithms/metric_short}

\subsection{Algorithm}
Algorithm~\ref{alg:metric} provides pseudocode of our metric for the evaluation of a single sequence.
Inputs are the ground-truth future $\mathbf{T}_\mathrm{gt}$, the multi-modal predictions $\mathbf{T}_{\mathrm{pred}}$ and the lane graph $\mathcal{G}$.
Output is a list $\mathbf{x}$ that contains the labels $1$ (miss) or $0$ (hit) per predicted mode.

At a high level, the algorithm works as follows:
The ground-truth endpoint gets assigned to a lane segment, more precisely to a point on the centerline of a lane segment (Line~\ref{alg:metric:gt_matching}).
Each predicted trajectory similarly gets assigned to lane segments (Line~\ref{alg:metric:pred_matching}).
For the predictions, multiple assignments are possible if multiple lane segments have a similar assignment confidence.
This allows the handling of overlapping lane segments.
If the distance along $\mathcal{G}$ from the assigned point of the ground-truth to any of the assigned points of the prediction is smaller than a velocity-dependent threshold $s_\mathrm{hit}$, the prediction is a hit (label $0$).
Otherwise, it is a miss (label $1$).

There are two special cases that are not defined in Algorithm~\ref{alg:metric}:
If there is no valid lane assignment for the ground-truth (e.g., ground-truth is outside of the drivable space), we fall back to a Euclidean MR and determine whether a prediction is a hit or miss by checking whether the predicted endpoint is within a radius of $s_\mathrm{hit}$ to the ground-truth endpoint or not.
If there is a valid lane assignment for the ground-truth but not for a predicted trajectory, the prediction is always a miss (label $1$).

\input{algorithms/metric_call}

In order to calculate LMR, misses are accumulated over the dataset $\mathcal{D}$.
Algorithm~\ref{alg:metric_call} provides pseudocode for this accumulation, which follows the principle of the original Euclidean MR.
LMR$_{@1}$ is defined as the ratio of sequences where the $k=1$ mode (mode with the highest confidence) is labeled as a miss.
LMR$_{@k}$ is defined as the ratio of sequences where all $k$ modes are labeled as a miss.

\subsection{Implementation Details} \label{subsec:implementation_details}
We implement our metric in Python.
The implementation is parallelized on a sequence-level, meaning that Line~\ref{alg:metric_call:get_miss_call} in Algorithm~\ref{alg:metric_call} gets executed multiple times in parallel.

In order to allow an efficient assignment of trajectories to centerlines of lane segments, an R-tree is used to query lane segments around a given point (Algorithm~\ref{alg:metric}, Line~\ref{alg:metric:r_tree}).
Therefore, lane centerlines are represented with minimum bounding rectangles.
This enables an efficient querying of centerlines' rectangles that intersect with the region around a defined point.

Finding valid paths along the lane graph $\mathcal{G}$ (Algorithm~\ref{alg:metric}, Line~\ref{alg:metric:dfs}) is done using a depth-first search.
We only allow a traversal via succeeding and preceding lane segments.

The values of $c_\mathrm{scale}$ and $c_\mathrm{const}$ are oriented on the original MR, which has a fixed Euclidean threshold of \SI{2}{m}.
They are chosen so that $s_\mathrm{hit} \approx \SI{2}{m}$ for $v_\mathrm{gt} = \SI{6.67}{m/s}$, which corresponds to the average agent velocity in the validation split of the dataset described in Section \ref{subsec:dataset}.

\section{Experiments}
In this section, we use our proposed metric for the evaluation of trajectory prediction models.
This includes the discussion of quantitative and qualitative results.

\subsection{Dataset} \label{subsec:dataset}
Experiments are carried out on the large-scale Argoverse~$2$ Motion Forecasting Dataset~\cite{Wilson2021}, containing \SI{199908}{} sequences for training, \SI{24988}{} for validation and \SI{24984}{} for testing.
Each sequence contains a High Definition (HD) map and the tracked states of agents in the surrounding of an autonomous vehicle.
These states are provided over a duration of \SI{11}{s} and sampled with \SI{10}{Hz}.
Given the initial \SI{5}{s} of a sequence, the goal is to predict the trajectory of the remaining \SI{6}{s} of one selected agent, namely the focal agent.

\subsection{Metrics}
To allow a comparison of our newly introduced lane distance-based metric, we additionally use the most established Euclidean distance-based metrics for the evaluation of trajectory prediction models, namely minADE, minFDE and MR.
For MR, the most common definition with a fixed threshold of \SI{2}{m} is used.
Evaluation is done for single- ($k=1$) and multi-modal ($k=6$) prediction.
We refer to Section~\ref{subsec:accuracy-based_metrics} for a detailed explanation of these metrics.

The evaluation is limited to vehicle-like agents only, meaning that only agents of type vehicle, motorcyclist and bus are considered.

\subsection{Trajectory Prediction Models}
We reimplement three different state-of-the-art trajectory prediction models on Argoverse~$2$ and benchmark them in order to compare our proposed LMR with the most widely used Euclidean distance-based metrics.
All models provide publicly available code for Argoverse~$1$.
Argoverse~$2$, however, is not limited to vehicles only, which is why we add a timestep-wise one-hot encoding of each agent's class to the input of all models.
The training protocols are kept the same as in the original publications.

\textbf{CRAT-Pred}~\cite{Schmidt2022} is a map-free prediction model that achieves state-of-the-art performance without making use of map information.
Agent trajectories are encoded via a long short-term memory.
Graph convolution and self-attention are used to aggregate temporal agent information with information of surrounding agents, namely the social context.
Linear residual decoders are used to predict trajectories.

\textbf{HiVT}~\cite{Zhou2022} consists of a local and a global encoder.
The local encoder first uses timestep-wise self-attention and a subsequent temporal transformer to encode the trajectory of an agent under consideration of its surrounding agents.
These local encodings are then fused with information of the HD map.
Local encoding is done for each agent in a scene separately.
A subsequent global encoder allows for an information exchange between the encodings resulting from the local encoder.
Multilayer perceptrons are used to predict trajectories.
Due to the computational complexity of the timestep-wise encoding of each agent's trajectory, including its social context given by surrounding agents, we limit the motion history to the last \SI{2}{s}.
This should have little impact on the results, as prior work in the field of learning-based pedestrian trajectory prediction shows that the deprivation of a long motion history does not lead to prediction degradation~\cite{Schoeller2020}.

\textbf{LaneGCN}~\cite{Liang2020} represents information of the HD map in a lane graph and uses graph convolution to extract information from this lane graph.
Agent trajectories are encoded via 1D-convolution neural networks and feature pyramid networks.
Four fusion cycles are used to fuse this information with information of the lane graph.
Predictions are conducted with linear residual decoders.

\subsection{Quantitative Results}
\input{tables/results}

\begin{figure*}[!t]
	\centering
	\subfloat[Sequences that are a miss in terms of our Lane Miss Rate but not in terms of the Euclidean Miss Rate]{%
		\includegraphics[width=0.33\textwidth,trim={3.5cm 3cm 5.9cm 3cm},clip,frame]{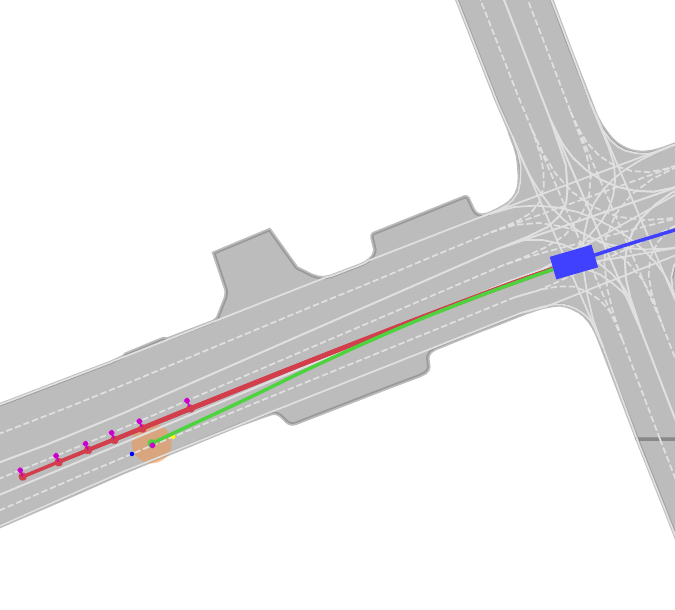}
		\hfill
		\includegraphics[width=0.33\textwidth,trim={3.19cm 3.812cm 3.79cm 0cm},clip,frame]{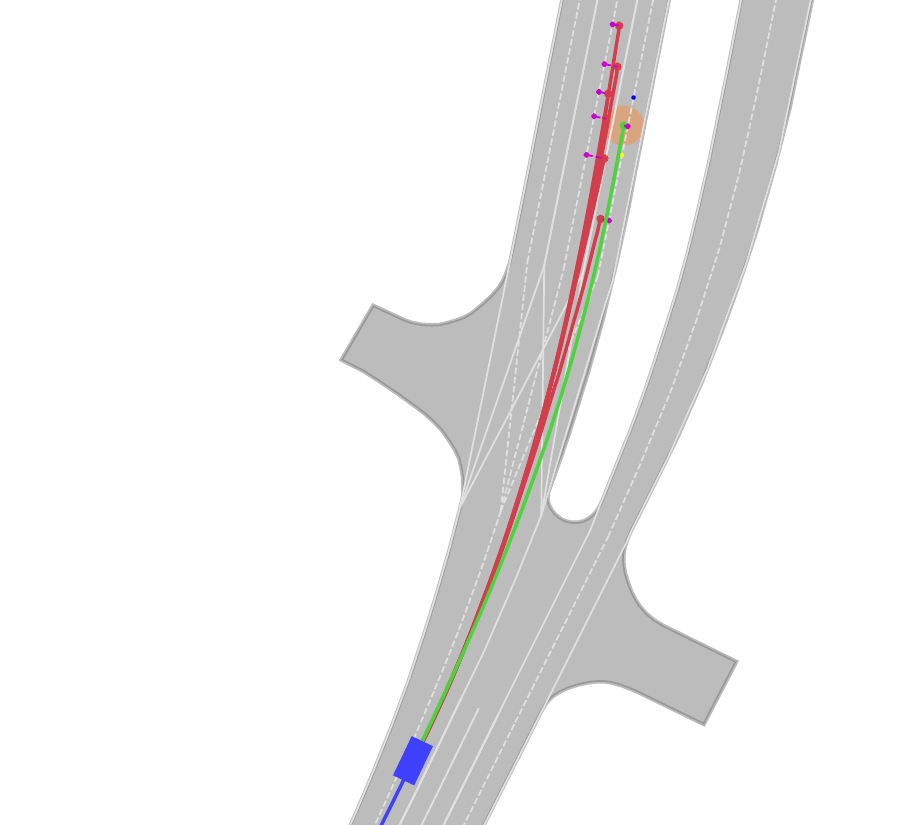}
		\hfill
		\includegraphics[width=0.33\textwidth,trim={3.86cm 1.602cm 4.66cm 3.6cm},clip,frame]{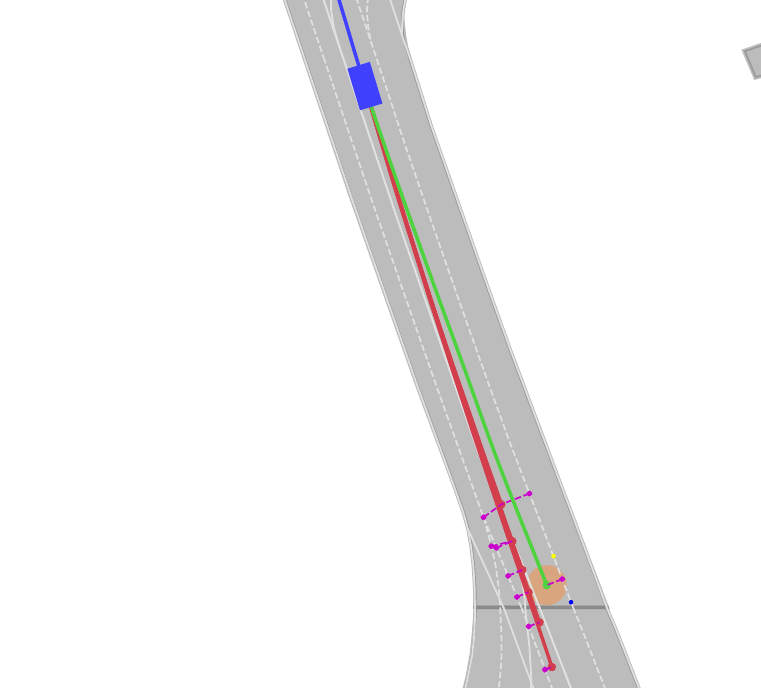}
		\label{subfig:euclideanhit_lanemiss}
	}
	\\
	\centering
	\subfloat[Sequences that are a miss in terms of the Euclidean Miss Rate but not in terms of our Lane Miss Rate]{%
		\includegraphics[width=0.33\textwidth,trim={4.46cm 4.8cm 5.26cm 1.5cm},clip,frame]{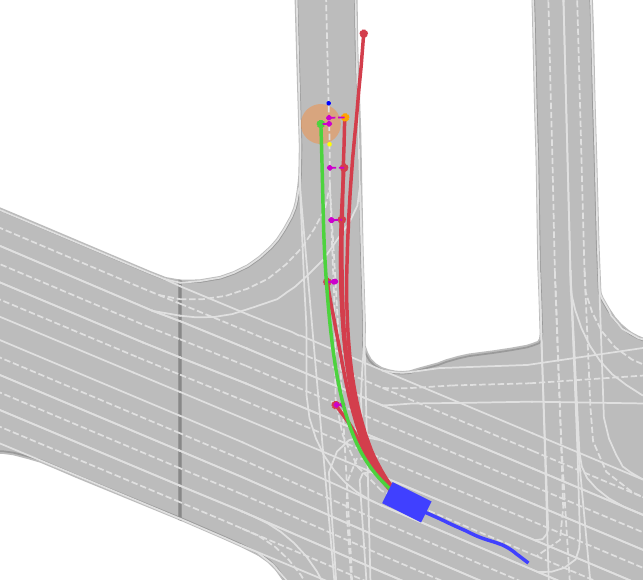}
		\hfill
		\includegraphics[width=0.33\textwidth,trim={5.7cm 2.4cm 2cm 2.08cm},clip,frame]{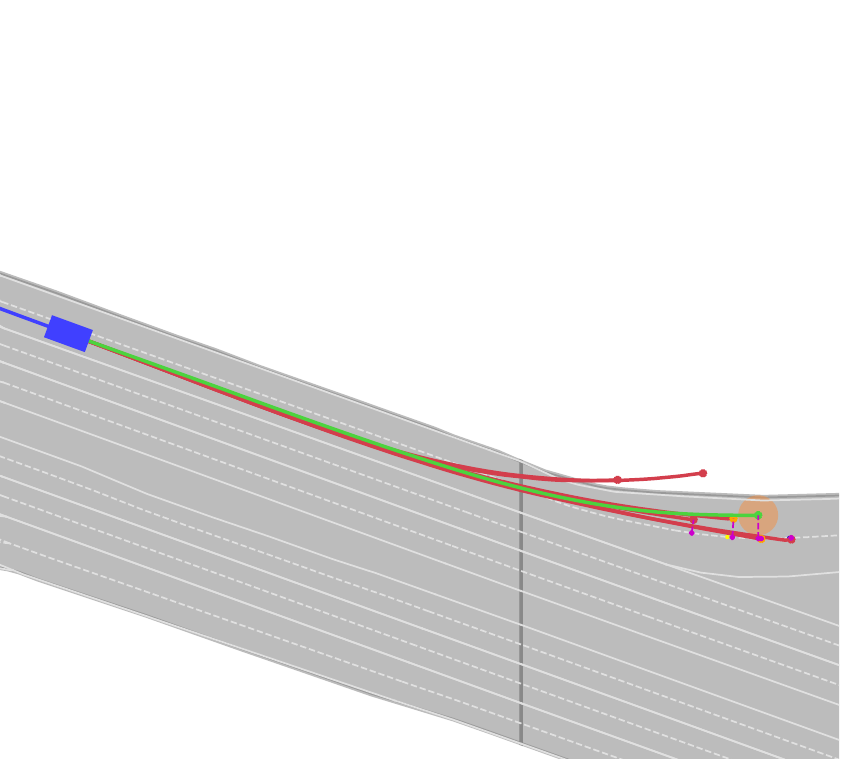}
		\hfill
		\includegraphics[width=0.33\textwidth,trim={5.135cm 4.296cm 6.335cm 3.58cm},clip,frame]{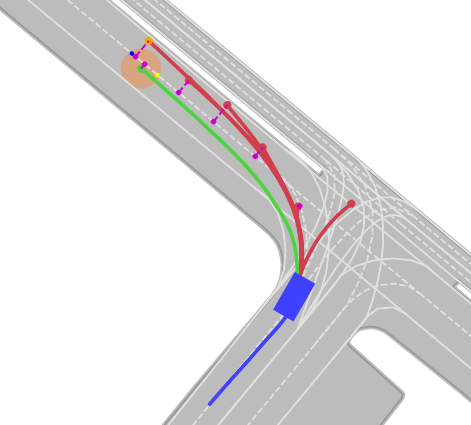}
		\label{subfig:euclideanmiss_lanehit}
	}
	\caption{Qualitative results of our Lane Miss Rate with LaneGCN: The past observed trajectory of the focal agent is colored in blue, the ground-truth future trajectory in green. Predictions are colored in red. Endpoints of ground-truth and predictions are marked with a dot. The assignments of the ground-truth and the predicted trajectories to the centerlines of lane segments are colored in purple. Predictions with an assignment that falls in between the blue and yellow dot on the ground-truth centerline yield a hit and vice versa. Predicted endpoints that yield a hit according to this definition are colored in orange instead of red. For comparison, the definition of the Euclidean Miss Rate is colored in light orange and centered around the ground-truth endpoint.}
	\label{fig:qualitative_results}
\end{figure*}

Table~\ref{tab:results} shows quantitative results on the Argoverse~$2$ validation split.
As described in Section~\ref{subsec:implementation_details}, thresholds were chosen to make LMR comparable to MR.
E.g., for CRAT-Pred the MR$_{@1}$ is $75.69 \%$ and the LMR$_{@1}$ is $74.27 \%$.
According to the Euclidean distance-based metrics, HiVT performs better with an MR$_{@1}$ of $71.82 \%$ and LaneGCN even more with an MR$_{@1}$ of $71.14 \%$.
The same holds for our lane distance-based metric with an LMR$_{@1}$ of $70.21 \%$ for HiVT and an LMR$_{@1}$ of $68.63 \%$ for LaneGCN.
An interesting observation is that LMR preserves the order of the Euclidean distance-based metrics in all quantitative results.

Another aspect is that the relative deviations between models differ for LMR compared to the other metrics.
For instance, the MR$_{@6}$ is $29.97 \%$ for HiVT and $25.87 \%$ for LaneGCN, which is a relative increase of $15.85 \%$.
However, the LMR$_{@6}$ is $33.30 \%$ for HiVT and $29.72 \%$ for LaneGCN, which is a relative increase of only $12.05 \%$.
The values indicate a lower effective difference in performance between HiVT and LaneGCN, which could only be identified with our new LMR metric.
Also, this confirms the benefit gained from an additional metric that uses lane distance instead of Euclidean distance.
We suggest investigating the performance of trajectory prediction models that use a lane-based loss, e.g.,~\cite{Wang2022}.
Additionally, we suggest investigating encoders that use information beyond lane segments, e.g.,~\cite{Monninger2023}, and see how this implicitly influences our lane distance-based metric.
Due to the lack of publicly available code, this is beyond the scope of this paper.

\subsection{Qualitative Results}
Fig.~\ref{fig:qualitative_results} shows qualitative examples where the definition of our LMR$_{@6}$ leads to different results than the standard Euclidean MR$_{@6}$.

Fig.~\ref{fig:qualitative_results} \subref{subfig:euclideanhit_lanemiss} shows sequences that yield a miss for all predicted trajectories in terms of LMR but not in terms of MR.
In all three sequences, the predictions do not cover the lane of the ground-truth, hence LMR results in a miss.
This illustrates that LMR implicitly weights Euclidean error relative to the lanes, by assigning ground-truth and predictions to centerlines of lane segments.
MR, however, is dependent on Euclidean distance and therefore weights error equally in all directions.
This results in the predictions not resulting in a miss, despite being located on the wrong lane.

Fig.~\ref{fig:qualitative_results} \subref{subfig:euclideanmiss_lanehit} shows sequences that yield a miss for all predicted trajectories in terms of MR but not in terms of LMR.
All three sequences show a clear trend:
The predictions are located on the correct lane but are outside the \SI{2}{m} MR radius.
Again, this illustrates that LMR implicitly weights Euclidean error relative to the lanes.
In this case, predictions are not labeled as a miss, because they are located on the correct lane and therefore cover the intent of the predicted vehicle.

\section{Conclusion}
We propose LMR, a novel lane distance-based metric for the evaluation of trajectory prediction models.
Quantitative and qualitative results on three state-of-the-art trajectory prediction models confirm the validity of this additional metric.
The benefit of LMR is that components of the Euclidean error are weighted relative to the underlying lanes.
This way, LMR extends the set of existing purely geometric metrics and goes into the direction of capturing intents of traffic agents.

Our publicly available source code of LMR for Argoverse~$2$ enables other researchers to evaluate their approaches with this lane distance-based metric.

\bibliographystyle{IEEEtran}
\balance
\bibliography{Literature}
\end{document}

%% file: algorithms/metric_short.tex
\begin{algorithm*}[!t]
	\small
	\caption{Lane miss calculation for multi-modal predictions}
	\label{alg:metric}
	\begin{algorithmic}[1]
	  	\Input
		\Desc{$\mathbf{T}_\mathrm{gt}$}{Ground-truth future trajectory}
		\Desc{$\mathbf{T}_{\mathrm{pred}}$}{Predicted future trajectories}
		\Desc{$\mathcal{G}$}{Lane graph}
		\EndInput
		\Output
		\Desc{$\mathbf{x}$}{Is miss}
		\EndOutput

		\Procedure{get\_is\_miss}{$\mathbf{T}_\mathrm{gt}, \mathbf{T}_{\mathrm{pred}}, \mathcal{G}$}
			\State Calculate average ground-truth velocity: $v_\mathrm{gt}$
			\State Calculate velocity-dependent lane hit threshold: $s_\mathrm{hit} \leftarrow (c_\mathrm{scale} \cdot v_\mathrm{gt}) + c_\mathrm{const}$	\Comment{We use $c_\mathrm{scale} = \SI{0.2}{s}$ and $c_\mathrm{const} = \SI{0.7}{m}$}
			\State Calculate most probable lane assignment of $\mathbf{T}_\mathrm{gt}$: $l_\mathrm{gt}, s_\mathrm{gt}, p_\mathrm{gt} \leftarrow \max\limits_p \left( \Call{get\_lane\_assignments}{\mathbf{T}_\mathrm{gt}, \mathcal{G}} \right)$ \label{alg:metric:gt_matching}
			
			\State Initialize empty miss list: $\mathbf{x} \leftarrow [\ ]$
			\For{$\mathbf{T}_{\mathrm{pred}, i}$ in $\mathbf{T}_{\mathrm{pred}}$}
				\State Calculate all $N$ lane assignments of $\mathbf{T}_{\mathrm{pred}, i}$: 
				\Statex[3]
					$[(l_{\mathrm{pred}, i, 1}, s_{\mathrm{pred}, i, 1},  p_\mathrm{\mathrm{pred}, i, 1}), \dots,  (l_{\mathrm{pred}, i, N}, s_{\mathrm{pred}, i, N},  p_\mathrm{\mathrm{pred}, i, N})] \leftarrow \Call{get\_lane\_assignments}{\mathbf{T}_{\mathrm{pred}, i}, \mathcal{G}}$ \label{alg:metric:pred_matching}
				\Statex[3] \Comment{We only use assignments where $p_\mathrm{\mathrm{pred}, i, n}$ is at most $0.1$ smaller than $\max (p_\mathrm{\mathrm{pred}, i, n})$}
				\State\algorithmicif{} {(Distance along $\mathcal{G}$ from lane segment $l_\mathrm{gt}$ at value $s_\mathrm{gt}$ to any lane segment $l_{\mathrm{pred}, i, n}$ at value $s_{\mathrm{pred}, i, n}$) $< s_\mathrm{hit}$} \algorithmicthen \label{alg:metric:dfs}
					\State \hspace{\algorithmicindent} $\mathbf{x}$.insert($0$) \algorithmicelse{} $\mathbf{x}$.insert($1$) \unskip\ \algorithmicend\ \algorithmicif
			\EndFor
			\State \Return $\mathbf{x}$
		\EndProcedure
		\\
		\Procedure{get\_lane\_assignments}{$\mathbf{T}, \mathcal{G}$}
			\State Initialize empty assignment list: $\mathbf{a} \leftarrow [\ ]$
			\For{Lane segment $l$ in $\mathcal{G}$} \label{alg:metric:r_tree}
				\If{$\boldsymbol{\tau}^{T_f}$ in lane boundaries of $l$}
					\State Calculate the shortest distance $d$ from $\boldsymbol{\tau}^{T_f}$ to the centerline of $l$
					\State Calculate assigned point $s$ on lane segment $l$ \Comment{Similiar to Frenet $s$}
					\State Calculate orientation $\alpha_\mathrm{traj}$ of trajectory at $\boldsymbol{\tau}^{T_f}$ and orientation $\alpha_l$ of lane segment at $s$
					\State Calculate orientation difference $\Delta \alpha \leftarrow |\mathrm{atan2}\left( \mathrm{sin}( \alpha_\mathrm{traj} - \alpha_l), \mathrm{cos}( \alpha_\mathrm{traj} - \alpha_l) \right)|$
					\State Calculate orientation assignment confidence $p_\alpha \leftarrow \mathrm{max} \left( 0, 1 - \frac{\Delta \alpha}{c_\mathrm{orient}} \right)$ 		\Comment{We use $c_\mathrm{orient} = \pi$}
					\State Calculate distance assignment confidence $p_d \leftarrow \mathrm{max} \left( 0, 1 - \frac{d}{c_\mathrm{dist}} \right)$ 	\Comment{We use $c_\mathrm{dist} = \SI{5}{m}$ \cite{Schmidt2022a}}
					\State Calculate combined assignment confidence $p \leftarrow w \cdot p_d + (1-w) \cdot p_\alpha$ 		\Comment{We use $w = 0.5$ (equal weighting)}
					\State $\mathbf{a}$.insert($(l, s, p)$)
				\EndIf
			\EndFor
			\State \Return $\mathbf{a}$
		\EndProcedure	
	\end{algorithmic}
\end{algorithm*}

%% file: algorithms/metric_call.tex
\begin{figure}[!t]
\vspace{-0.3cm}
\begin{algorithm}[H]
	\small
	\caption{Lane Miss Rate calculation over a full dataset}
	\label{alg:metric_call}
	\begin{algorithmic}[1]
	  	\Input
		\Desc{$\mathcal{D}$}{Dataset, $|\mathcal{D}| = d$}
		\EndInput
		\Output
		\Desc{$\mathrm{LMR}$}{Lane Miss Rate}
		\EndOutput

		\Procedure{get\_lane\_miss\_rate}{$\mathcal{D}$}		
			\State Initialize empty miss list: $\mathbf{X} \leftarrow [\ ]$
			\For{($\mathbf{T}_\mathrm{gt}, \mathbf{T}_{\mathrm{pred}}, \mathcal{G}$) in $\mathcal{D}$}
				\State $\mathbf{X}$.insert(\Call{get\_is\_miss}{$\mathbf{T}_\mathrm{gt}, \mathbf{T}_{\mathrm{pred}}, \mathcal{G}$}) \label{alg:metric_call:get_miss_call}
			\EndFor
			\State Convert list of lists to 2D-array $\mathbf{X} \leftarrow$ \Call{to\_array}{$\mathbf{X}$},
			\Statex[2] $\mathbf{X} \in \mathcal{R}^{d \times k}$
			\State $\mathrm{LMR}_{@1} = \frac{\mathrm{sum(\mathbf{X}[:, 0])}}{d}$
			\State $\mathrm{LMR}_{@k} = \frac{\mathrm{sum(all(\mathbf{X}, axis=1))}}{d}$
			\State \Return $\mathrm{LMR}$
		\EndProcedure
	\end{algorithmic}
\end{algorithm}
\vspace{-0.8cm}
\end{figure}

%% file: tables/results.tex
\begin{table*}[!t]
	\caption{Quantitative results on the Argoverse~$2$ validation split}
	\label{tab:results}
	\setlength{\tabcolsep}{3.6pt}
	\centering
	\begin{tabularx}{0.65\linewidth}{Xllllllll}
		\toprule
		\multirow{2}{*}{Model} &            \multicolumn{4}{c}{$k=1$}            &      \multicolumn{4}{c}{$k=6$}      \\
		                       & minADE & minFDE & MR      & LMR  \hspace{0.6cm} & minADE & minFDE & MR      & LMR     \\ \midrule
		CRAT-Pred              & $2.44$ & $6.40$ & $75.69$ & $74.27$             & $1.29$ & $2.76$ & $42.60$ & $44.29$ \\
		HiVT                   & $2.31$ & $6.23$ & $71.82$ & $70.21$             & $0.97$ & $1.99$ & $29.97$ & $33.30$ \\
		LaneGCN                & $2.22$ & $6.00$ & $71.14$ & $68.63$             & $0.90$ & $1.71$ & $25.87$ & $29.72$ \\ \bottomrule
	\end{tabularx}
	\vspace{0.2cm}
\end{table*}